\documentclass[letterpaper]{article} % DO NOT CHANGE THIS
\usepackage{aaai24}  % DO NOT CHANGE THIS
\usepackage{times}  % DO NOT CHANGE THIS
\usepackage{helvet}  % DO NOT CHANGE THIS
\usepackage{courier}  % DO NOT CHANGE THIS
\usepackage[hyphens]{url}  % DO NOT CHANGE THIS
\usepackage{graphicx} % DO NOT CHANGE THIS
\usepackage{natbib}  % DO NOT CHANGE THIS AND DO NOT ADD ANY OPTIONS TO IT
\usepackage{caption} % DO NOT CHANGE THIS AND DO NOT ADD ANY OPTIONS TO IT
\usepackage{algorithm}
\usepackage{algorithmic}
\usepackage[title]{appendix}

\usepackage{listings}
\DeclareCaptionStyle{ruled}{labelfont=normalfont,labelsep=colon,strut=off} % DO NOT CHANGE THIS
\floatstyle{ruled}
\newfloat{listing}{tb}{lst}{}
\floatname{listing}{Listing}
\usepackage{multirow}
\usepackage{graphicx}
\usepackage{booktabs}
\title{
F$^{3}$-Pruning: A Training-\underline{F}ree and Generalized Pruning Strategy \\
towards \underline{F}aster and \underline{F}iner Text-to-Video Synthesis}

\author{
    Sitong Su\equalcontrib,
    Jianzhi Liu\equalcontrib,
    Lianli Gao,
    Jingkuan Song
}
\affiliations{
    University of Electronic Science and Technology of China (UESTC)\\
    sitongsu9796@gmail.com
}

\usepackage{bibentry}
\begin{document}

\twocolumn[{%
\renewcommand\twocolumn[1][]{#1}%
\maketitle
\begin{center}
    \centering
    \includegraphics[width=0.9\linewidth]{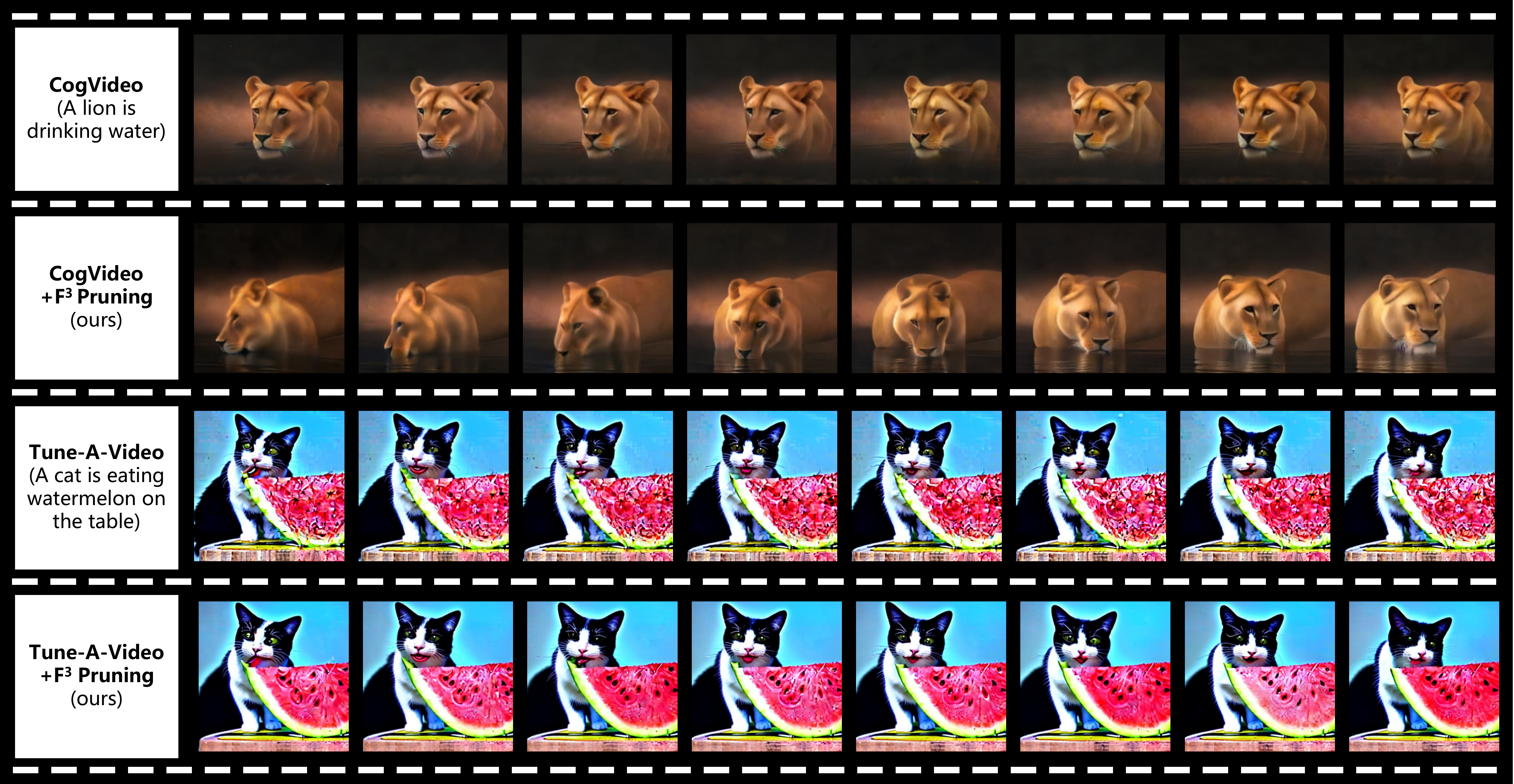}
    \captionof{figure}{\textit{Demonstration of some visual results comparison in Text-to-Video (T2V) synthesis}. Our F$^3$-Pruning is applied to the classic transformer-based method CogVideo and the typical diffusion-based method Tune-A-Video. \textbf{Without any extra training}, F$^3$-Pruning not only boosts inference efficiency of T2V but also enhances video quality. On the public video dataset UCF-101, applying F$^3$-Pruning to CogVideo makes it \textbf{1.35x} faster and promotes video quality metrics FVD by \textbf{22\%}.}
    \label{banner}
\end{center}%
}]

\begin{abstract}
Recently Text-to-Video (T2V) synthesis has undergone a breakthrough by training transformers or diffusion models on large-scale datasets.
Nevertheless, inferring such large models incurs huge costs.
Previous inference acceleration works either require costly retraining or are model-specific.
To address this issue, instead of retraining we explore the inference process of two mainstream T2V models using transformers and diffusion models.
The exploration reveals the redundancy in temporal attention modules of  both models, which are commonly utilized to establish temporal relations among frames.
Consequently, we propose a training-free and generalized pruning strategy called \textbf{F$^3$-Pruning} to prune redundant temporal attention weights.
Specifically, when aggregate temporal attention values are ranked below a certain ratio, corresponding weights will be pruned.1
Extensive experiments on three datasets using a classic transformer-based model CogVideo and a typical diffusion-based model Tune-A-Video verify the effectiveness of F$^3$-Pruning in inference acceleration, quality assurance and broad applicability.

\end{abstract}

%%%%%%%%%%%%%%%%%%%%%%%%%%%%%%%%%%%%%%%%%%%%%%%%%%%%%%
\section{Introductions}
Text-to-Video (T2V) synthesis aims to generate high-quality and temporally coherent videos semantically aligned with text inputs.
In the early stage of T2V, traditional generative models like GAN~\cite{gan} are trained on datasets of limited scale. Hence, the synthesized videos are restricted by limited semantics and low quality~\cite{LiMSCC18, PanQYLM17}.
To tackle this challenge, recent T2V models utilize large and powerful generative models transformer or diffusion models to train on large-scale datasets~\cite{HoSGC0F22,WuLJYFJD22}.
Alternately, other recent T2V models~\cite{HongDZLT23,HoCSWGGKPNFS22,WuGWLGHSQS22} build upon large-scale T2I models~\cite{cogview, imagen, latentdiffusion} by introducing temporal information, which are also based on transformers or diffusion models.
Despite extraordinary progress in video quality and zero-shot generalization ability, inferring such large models incurs huge costs.

Nevertheless, previous inference acceleration methods for transformer or diffusion models either incorporate costly retraining or cannot be generally applied to both two models.
Specifically, for transformers-based models, several works are dedicated towards a more efficient transformer~\cite{dosovitskiy2020image, mehta2021mobilevit, meng2022adavit}. Whereas, these works focus on discriminative tasks without considering synthesis quality.
Other works like~\cite{MaoXNBFLXX21, WeiTSYJ23, RaoZLLZH21} prune weights or tokens with costly retraining or finetuning. Differently, ToMe merges tokens without training, which however introduces time-consuming manipulation in inference. For diffusion-based models, traditional diffusion acceleration speeds up the denoising process~\cite{lu2022dpm}, which is limited by the trade-off between video quality and efficiency. Other methods like knowledge distillation~\cite{HoCSWGGKPNFS22} or diffusion model pruning~\cite{FangMW23} also incorporate extra training costs. All of the above works are specially designed for one specific model, thus hampering their applicability to different T2V models.

\begin{figure*}[t]
	\centering
	\includegraphics[width=1\textwidth]{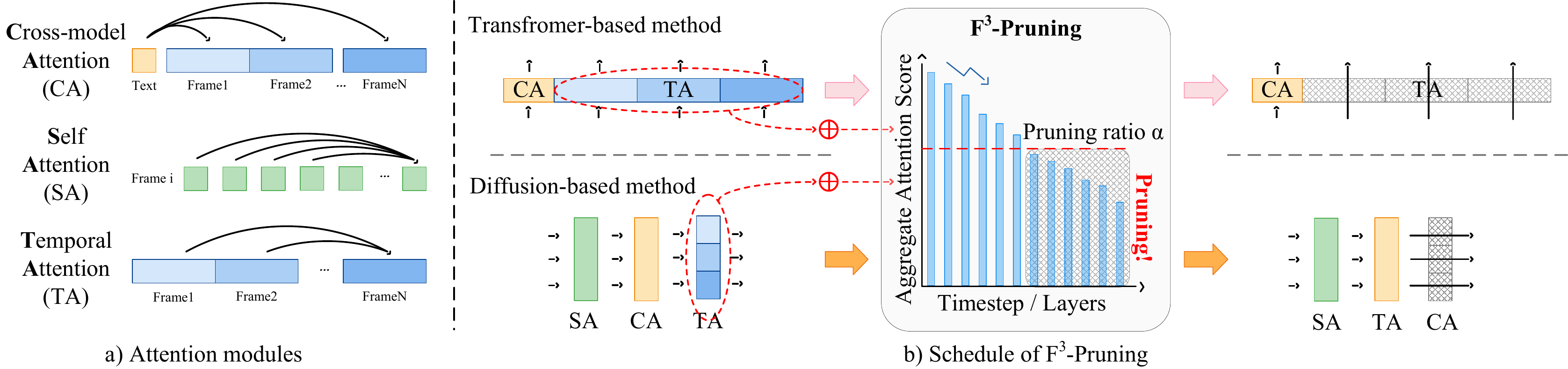}
	\caption{\textit{Overview of our proposed F$^3$-Pruning}. In a), we show three attention modules Cross-model Attention (CA), Self Attention (SA) and Temporal Attention (TA), which are commonly used in T2V to respectively model text-visual alignment, visual quality within each frame and temporal coherence among frames. In b), we demonstrate the schedule of our F$^3$-Pruning applied to the transformer-based methods and the diffusion-based methods. TA weights will be pruned when the sums of TA values of some network layers or denoising timesteps, called Aggregate Attention Score, are ranked below a pruning ratio $\alpha$.}
	\label{overview}
\end{figure*}

% To avoid retraining and increase generalization, 
To address the issue, instead of retraining we investigate the inference process of a classic transformer-based model CogVideo~\cite{HongDZLT23} and a typical diffusion-based model Tune-A-Video~\cite{WuGWLGHSQS22} to search for the common points.
Regardless of model types, the above two models both establish attention between text to each frame, within frames and cross frames to respectively model text-visual alignment, visual quality and temporal coherence as shown in Fig.~\ref{overview}(a). We respectively name these three modules as Cross-modal Attention(CA), Self Attention(SA) and Temporal Attention(TA). Note that, CA, SA and TA are widely adopted in recent T2V models.

Therefore, we statistically analyze the attention values distribution of the three modules during the inference process as shown in Fig.~\ref{AAS}. At the bottom of Fig.~\ref{AAS}, there are attention maps visualization respectively for CogVideo and Tune-A-Video. The diagonal line of the map refers to SA while the other parts represent TA. As observed, TA is full of values approaching zero, which indicates there exist plenty of non-contributory and redundant parts getting involved in temporal modeling. 
The redundancy can also be verified by visual results in the first row of Fig.~\ref{banner} that the motion of the lion is restricted.
Moreover, the histograms in Fig.~\ref{AAS} demonstrate the relationship between summed attention values called Aggregate Attention Score(AAS) and network layers or denoising timesteps.
As shown, the AAS of TA keeps declining while that of CA or SA increases with the generation process goes.
It implies that the importance of generation is gradually transferred from temporal information to visual quality and text-visual alignment.
And TA plays a less important role in the late stage of generation.
In summary, in the inference stage of both transformers and diffusion models, there exists redundancy in temporal attention modules.

Inspired by the above observation, we propose a training-free and generalized pruning strategy called F$^3$-Pruning to prune redundant and less important temporal attention weights as shown in Fig.~\ref{overview}(b). Instead of designing intrinsic pruning criteria, we claim that attention values could reasonably represent the saliency of corresponding attention weights. Besides, considering the sparsity of TA values, TA values are aggregated into Aggregate Attention Scores (AAS) as pruning criteria rather than complicated comparisons. As demonstrated in Fig.~\ref{overview}(b), our F$^3$-Pruning is applied over network layers or denoising timesteps with the pruning criteria to cut off TA weights whose AAS are ranked below a specific ratio.
With F$^3$-Pruning, redundant TA is released and redistributed to SA or CA to further promote video quality and text-visual alignment with coherence.

Our contributions could be summarized as follows:

1) We explore the inference process of two mainstream T2V models using transformer and diffusion, which reveals the redundancy of temporal attention in both models.

2) We propose a training-free and generalized pruning strategy called F$^3$-Pruning to prune redundant temporal attention weights, which both speeds up T2V inference and assures video quality.

3) Extensive experiments on three datasets prove the effectiveness, efficacy and generalization of F$^3$-Pruning. Particularly, F$^3$-Pruning applied to CogVideo on UCF-101 dataset not only speeds up CogVideo by \textbf{1.35x} but also significantly improves video quality metric FVD by \textbf{22\%}.

%%%%%%%%%%%%%%%%%%%%%%%%%%%%%%%%%%%%%%%%%%%%%%%%%%%%%%
\section{Related Works}

\subsection{Text-to-Video Generation}
In the initial stage of development, some GAN-based~\cite{LiMSCC18, PanQYLM17} models can only create low-resolution videos with restricted semantics.
% because of the adversarial process. 
Subsequent works mainly focus on non-adversarial processes. 
A classic diffusion-based work VDM~\cite{HoSGC0F22}, naturally extends a standard image diffusion model and enables it jointly training with image and video data. Different from that, some works~\cite{WuLJYFJD22, VillegasBKM0SCK23, GeHYYPJHP22} modify a base transformer to adapt to the video synthesis involving multi-model signals.

However, these works require exceptionally high training costs for training from scratch. 
Therefore some researchers explore the utilization of pretrained T2I models to ease training costs for T2V generation.
For instance, CogVideo~\cite{HongDZLT23} devises a large-scale T2I transformer CogView2~\cite{DingZHT22} by introducing the temporal information through the use of attention among frames. On a different note, Make-A-Video~\cite{SingerPH00ZHYAG23} breaks away from the usual reliance on text-video pairs for T2V generation by utilizing a pretrained T2I model. 
%This eliminates the need for text-video pairs training. 
Imagen Video~\cite{HoCSWGGKPNFS22} follows Imagen~\cite{SahariaCSLWDGLA22}, employing a cascaded diffusion model with attention and convolution at multiple resolutions. Moreover, as the quality of video generation improves, recent works begin to explore diverse settings of generation. Tune-A-Video~\cite{WuGWLGHSQS22} proposes a one-shot video tuning approach for T2V generation, incorporating temporal attention into Stable diffusion~\cite{RombachBLEO22}. And Text2Video-Zero~\cite{KhachatryanMTHWNS23} enables zero-shot T2V generation without training video.

Since these methods primarily prioritize video quality, semantical consistency, and coherence, they overlook the issue of huge computational costs and long latency in reference.

\subsection{Inference Acceleration}
% Traning free and training or fine-tune 2 pra.
The methods for model acceleration at the algorithmic level are mainly divided into pruning~\cite{XiaXC22, LagunasCSR21, ZhuoWLW022}, quantization~\cite{QinDZYLDL022, LiYWC22, LiuOPXYLKM22, SongZGXLS19}.
In this work, we focus on pruning, which is classified into structural and unstructured pruning. Some works focus on unstructured pruning~\cite{DongCP17, ParkLMS20, Sanh0R20, LeeAGT20}, which finds the unimportant parameters and conceals them by masking or setting them to zero. Those methods usually do not bring actual acceleration. Whereas very recent works perform on structural pruning~\cite{DingDGH19, YouYYM019, LiuZKZXWCYLZ21}, which stands out due to its capability to physically remove parameters and substructures from neural networks for accelerating training or inference. Nevertheless, the above research on network pruning has predominantly concentrated on tasks involving discrimination, notably classification endeavors and these approaches all require additional training to determine the locations for pruning.

To avoid training costs, Bolya et al.~\cite{BolyaH23, BolyaFDZFH23} propose a token merging method that notably enhances inference speed by reducing tokens through cosine similarity, all without requiring any training. Nevertheless, this approach entails extra computations to determine which tokens should be merged. This can potentially add substantial overhead to the inference process. Our approach is different from prior methods as we harness temporal redundancy, negating the necessity for intricate token similarity comparisons and sidestepping supplementary computations during inference. In this study, we change the token-specific operations to layer-specific operations by analyzing temporal redundancy and demonstrate the feasibility of this pruning approach in video generation.

%%%%%%%%%%%%%%%%%%%%%%%%%%%%%%%%%%%%%%%%%%%%%%%%%%%%%%
\section{Methodology} 
\label{Methods}
Firstly, we introduce some preliminaries in T2V in Sec.~\ref{3_1} for further illustration. Then in Sec.~\ref{3_2}, we investigate the inference stage of two mainstream T2V models to analyze the redundant parts for pruning. Based on the analysis, we introduce our pruning strategy F$^3$-Pruning in Sec.~\ref{3_3}.

\subsection{Preliminary}
\label{3_1}
% Existing Text-to-Video synthesis models are mainly built upon large-scale Text-to-Image models by introducing temporal information, to fully utilize the strong capability of T2I models. 
Recent Text-to-Video models are mainly established on powerful generative models transformers and diffusion models.
% Considering generative model types, T2V models generally bifurcate into 1) transformer-based models and 2) diffusion-based models. 
Despite the model types, Cross-modal Attention (CA), Self Attention (SA) and Temporal Attention (TA) are commonly utilized to respectively assure text-visual alignment, visual quality of each single frame and temporal coherence among frames.
Specifically, as depicted in Fig.~\ref{overview} a), CA models attention between text features $T$ and visual features of each frame $F_i$. And SA establishes attention matrices within each frame for visual quality. Learning attention assignation among different frames forms TA for temporal modeling. Consequently, the formulations of these three attention modules could be summarized as follows:
\begin{equation}
\centering
\left\{
\begin{array}{cc}
     CA = Attn(T, F_i), &  \\
     SA = Attn(F_i, F_i), & i,j \in (1,...,N) \\
     TA = Attn(F_i, F_j),
\end{array}
\right.
\end{equation}
where $N$ refers to the number of generated video frames. When $i = j$, SA is contained in TA. According to the formulations, the time complexity of TA is $\mathcal{O}(N^2)$ in contrast to $\mathcal{O}(N)$ of SA, which indicates a quadratic time consumption with the increas of frames or resolution.

For the transformer-based methods, we consider CogVideo~\cite{HongDZLT23}, a classic large-scale open-source T2V model as an example.
As depicted on the left top of Fig.~\ref{overview} b), all video frames and text are discretized into tokens and different attentions are entangled to jointly model text and visual connections as 
\begin{equation}
\centering
[CA; TA],
\end{equation}
where SA is contained in TA.

For the diffusion-based methods, we select Tune-A-Video~\cite{WuGWLGHSQS22} which serves as a baseline one-shot T2V model.
As shown on the left bottom of Fig.~\ref{overview} b), in each denoising step, CA, SA and TA are disentangled and arranged in a cascaded way as
\begin{equation}
\label{tav_eq}
\centering
TA(CA(SA)).
\end{equation}

\begin{figure}[t]
	\centering
	\includegraphics[width=1\linewidth]{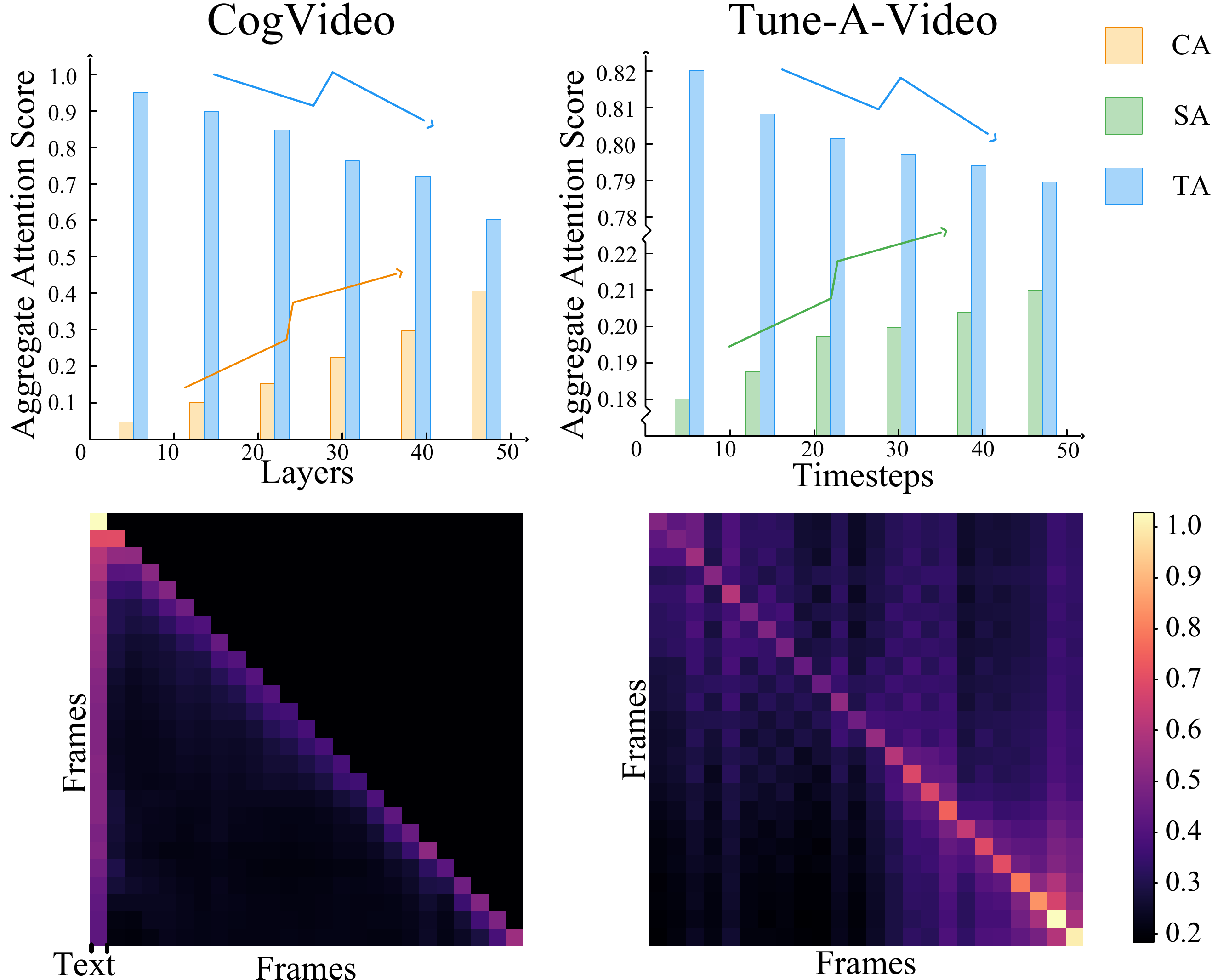}
	\caption{\textbf{Top}: Demonstration of the relation between Aggregate Attention Score (AAS) and network layers or denoising timesteps. AAS is declining with the inference step. \textbf{Bottom}: Attention Visualization. The diagonal line represents SA, and the upper and lower triangles represent TA. In particular, the leftmost bright line in CogVideo represents CA. As seen, attention values are sparsely distributed.
 % The trend of the sum of attention score called Aggregate Attention Score(AAS) with inference step (left) and attention matrix for visualization (right). 
 %On the right subfigure: diagonal represents Self Attention within frames, upper and lower triangular areas represent Cross Attention among frames. Notably, the leftmost vertical bright line denotes Text Attention in CogVideo.
 }
	\label{AAS}
\end{figure}

% \subsection{Analysis of Temporal Redundancy}
\subsection{What to Prune: Redundancy Analysis}
\label{3_2}
To avoid costly retraining or finetuning, we probe into the inference stage of the above two representative T2V models CogVideo and Tune-A-Video.
Since CA, SA and TA are commonly shared among T2V models, we choose to analyze their attention values distribution in inference, which are summarized into the following three observations.

\textbf{Observation 1.} \textit{A large portion of temporal attention values approaching zero indicates redundancy for temporal modeling.} For direct observation, we visualize attention maps of CogVideo and Tune-A-Video in the bottom of Fig.~\ref{AAS}. Specifically, for the attention map of CogVideo in the left bottom, values in the most left column, the diagonal line and the other parts in the lower triangular matrix respectively represent CA, SA and TA. As seen, attention values are intensively assigned to CA and SA while sparsely distributed to TA. Moreover, in TA the neighboring frames are more dedicated compared to more previous frames. The conclusion is also compatible with the common sense that a frame is mostly related to its neighboring frames or itself rather than attending to all other frames. In addition, for the attention map visualization of Tune-A-Video in the right bottom of Fig.~\ref{AAS}, the diagonal line refers to SA and the other parts refer to TA. Also, SA is much more emphasized compared to TA, which is consistent with the observation in CogVideo. 

Generally, the distribution of attention values indicates that analyzing linguistic information and building visual information within a frame matters much more than attending to all the precedent frames. Based on the observation above, we can conclude that there exists a large extent of redundancy for fully connecting each two frames in TA. The redundancy can also be verified by visual results of CogVideo in Fig.~\ref{banner}. The dense connection among frames restricts the motion of the video, where the lion is kept unchanged instead of drinking water.  

\textbf{Observation 2.} \textit{The decrease of aggregate temporal attention values with generation process implies its declining importance.}
For further analysis, we utilize text prompts from public datasets, which are fed into CogVideo and Tune-A-Video to infer corresponding videos. Then in different network layers or denoising timesteps, we statistically collect the sum of attention values in CA, SA and TA, which are called Aggregate Attention Score (AAS). The corresponding results form the histogram in the top of Fig.~\ref{AAS}. As observed, with the deepening network layers in the transformer or increasing denoising timesteps in the diffusion model, the AAS keeps decreasing in TA in contrast with the rise of CA or SA. This implies the decreasing importance of TA along with the generation process. 
For CogVideo, the increasing AAS in CA demonstrates that generation is more oriented towards textual information, which is consistent with the analysis in CogVideo. As for Tune-A-Video, as the denoising process goes, the focus of generation attends more and more to visual details within a frame rather than temporal relations among frames. The observation also corresponds to a conclusion of diffusion models that the early denoising process focuses on low-frequency contents while the late one focuses more on high-frequency details. 

The overall observations reveal that temporal modeling takes different proportions in the whole generation process, and matters less important in the late stage of generation.

\textbf{Observation 3.} \textit{Removing redundant TA leads to reasonable attention redistribution to CA or SA, thus contributing to video quality.} 
% As shown by in Fig.~\ref{overview}(b), CA, SA and TA are entangled or cascaded in mainstream T2V models. 
For CogVideo, TA and CA are entangled modeling as shown in the left top of Fig.~\ref{overview}(b), which means that their normalized attention values are summed to be 1. Once TA is released, its attention values will be redistributed to CA, thus contributing to text-visual alignment. For Tune-A-Video, CA, SA and TA are cascaded connected as described in Eq.~\ref{tav_eq}. When the overly tight restriction of TA is removed, the influence of SA and CA will be enlarged correspondingly. Especially, if less important TA weights in the late stage of the denoising process are pruned, focusing more on SA will favor the construction of high-frequency details. This observation can be verified through visual results of Tune-A-Video+F$^3$-Pruning as shown in Fig.~\ref{banner}.

In summary, observations 1 and 2 indicate the redundancy of TA and decreased importance of TA in the late stage of generation. Moreover, observation 3 further analyzes the merits of releasing redundant TA, which will contribute to video quality or text-visual alignment.

% Sparse Cross Attention and Declining Cross Attention motivate us to prune the redundant and less focused Cross Attention in the generation process.
\begin{table*}[]
\centering
\resizebox{0.85\linewidth}{!}{%
\begin{tabular}{@{}l|cc|cc|c|cc@{}}
\toprule
\multicolumn{1}{c|}{\multirow{2}{*}{Method}} & \multicolumn{2}{c|}{UCF-101}             & \multicolumn{2}{c|}{DAVIS}           & LONGTEXT             & \multicolumn{2}{c}{Performance}       \\
                        & FVD $\downarrow$ & Clip(text) $\uparrow$ & FVD $\downarrow$ & Clip(text) $\uparrow$ & Clip(text) $\uparrow$ & PFLOPS $\downarrow$   & Time (s) $\downarrow$ \\ \midrule
CogVideo                & 2537             & \textbf{25.03}        & 1851             & \textbf{26.34}        & 28.07                 & 9484          & 486                   \\ \midrule
+ Pruning (*)             & 4073             & 24.66                 & 2239             & 25.46                 & 27.94                 & \textbf{5271} & 672                   \\
+ Reorganization  (*)     & 4091             & 24.61                 & 2037             & 25.81                 & 27.73                 & 5280          & 688                   \\
+ TPS  (*)                & 4471             & 24.35                 & 2378             & 25.16                 & 27.83                 & 5280          & 704                   \\
+ ToMe  (*)               & 3594             & 24.74                 & 1837             & 26.34                 & 28.07                 & 5290          & 985                   \\
+ F$^3$-Pruning               & \textbf{1990}    & 24.77                 & \textbf{1781}    & 26.21                 & \textbf{28.4}         & 5311          & \textbf{359}          \\ \bottomrule
\end{tabular}
}
\caption{Quantitative comparison results on five pruning methods applied to the typical transformer-based model CogVideo.
% We measured the following metrics: Fréchet Video Distance (FVD), Clip-score, PFLOPS (PetaFLOPS, $10^{15}$ FLOPS), and inference time(s). 
``*" represents that the method itself is not originally designed for auto-regressive transformers, and we adapt it to fit CogVideo.}
\label{tab:compAttention}
\end{table*}

\subsection{How to Prune: F$^3$-Pruning}
\label{3_3}
Motivated by the observation and analysis from Sec.~\ref{3_2}, we propose F$^3$-Pruning to structurally prune weights of redundant temporal attention as shown in Fig.~\ref{Methods}(b). 
% And we will elaborate our algorithm in pruning criteria and pruning schedule as follows.

\textbf{Aggregate Attention Score.} Instead of designing complicated pruning criteria that usually demand extra heads to be trained, we claim that attention values could act as a reasonable indication of the corresponding weights importance.
Considering the sparsity of temporal attention, bits-by-bits attention values comparison or ranking will not only cause a large amount of calculations but also hurts the robustness due to the variance of comparing vectors with small norms. In contrast, comparing the sum of temporal attention values avoids dense matrix calculations and could aggregate the sparse values to better represent the importance of the overall temporal attention.
% And we hypothesize that the trend of the attention values in the same layer of different tokens remains the same.
Consequently, we propose to sum over the temporal attention values for each network layer or denoising timestep as our pruning criteria coined as Aggregate Attention Score (AAS). The corresponding formulation could be summarized as follows:
\begin{equation}
\label{aas}
\centering
AAS = sum(Attn(F_i, F_j)) \quad i \neq j.
\end{equation}

\textbf{F$^3$-Pruning.} Given the criteria, our F$^3$-Pruning schedule is illustrated in Fig.~\ref{Methods}(b).
Specifically, we iterate through all the training samples in a public video dataset and infer corresponding video samples using the given T2V model.
Afterward, we calculate over each network layer or denoising timestep following Eq.~\ref{aas} to obtain the AAS.
According to the hypothesis that `smaller-norm-less-important', we rank the obtained AAS for the whole generation process by their norms.
Then we prune the temporal attention weights with `smaller-norm' below a fixed threshold.
Note that different T2V models own different extents in temporal redundancy, which means a fixed threshold will not be adaptable. Therefore, we choose to prune a proportion of temporal attention weights defined by a pruning ratio $\alpha$, where $0 \leq \alpha \leq 1$. That is to say, we prune $\alpha$ of temporal attention weights whose AAS rank in the last $\alpha$ part. 
Regarding the decreasing importance of temporal attention, we actually prune the TA weights in the late stage of generation.

\begin{figure*}[]
	\centering
	\includegraphics[width=0.87\linewidth]{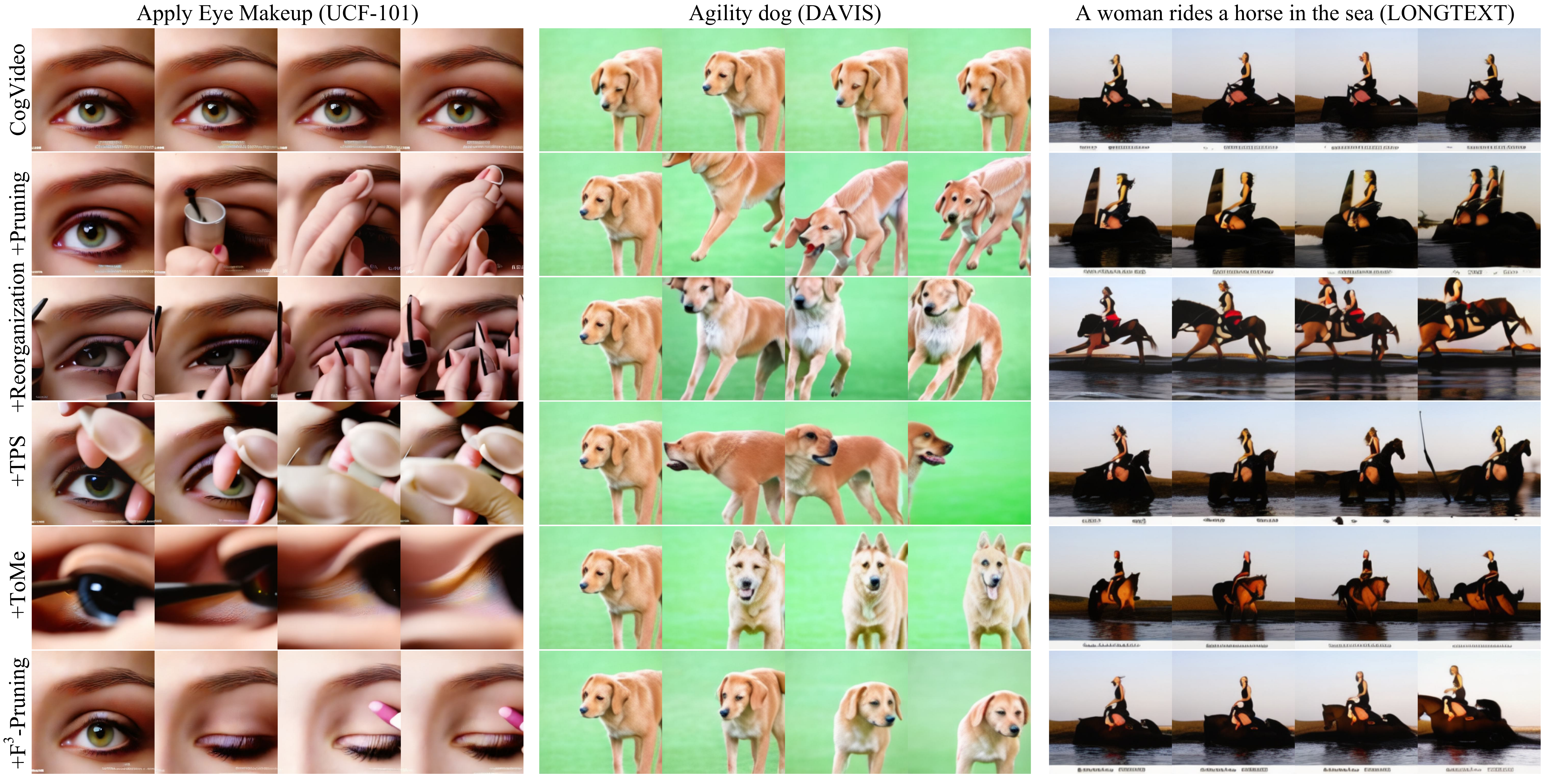}
	\caption{Some examples generated by five pruning methods applied to CogVideo. As demonstrated, F$^3$-Pruning performs the best in coherence and text comprehension.}
	\label{transformerComp}
\end{figure*}
\begin{figure*}[t]
	\centering
	\includegraphics[width=0.9\linewidth]{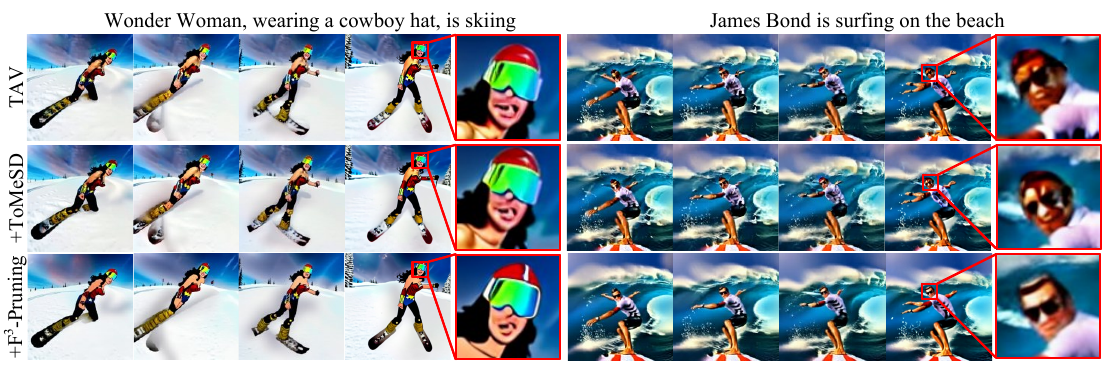}
	\caption{Some examples generated by two pruning methods applied to the Tune-A-Video (TAV) on the datasets of LONGTEXT. As demonstrated, F$^3$-Pruning performs the best, especially in object details.}
	\label{diffusionComp}
\end{figure*}

\section{Experiments and Analysis} \label{Experiments}

\subsection{Datasets and Evaluation Metrics}

\textbf{Datasets} To prove the effect of F$^3$-Pruning, and compare it to other SOTA pruning methods. 
% Our method is applicable to a wide range of text-guided video generation models. 
% Since the loss of text-video pairs,
We follow the settings of~\cite{HongDZLT23, WuGWLGHSQS22} and select two public video datasets, including a) \textbf{UCF-101}, an action recognition dataset~\cite{PengZZ19} and b) \textbf{DAVIS}, a densely annotated video segmentation~\cite{VoigtlaenderLTL20}, which both contain label-to-video pairs. Furthermore, to address the potential loss of text information from using labels as inputs, we build a c) \textbf{LONGTEXT} dataset, which consists of full sentences generated by ChatGPT\cite{openai_gpt}.

\textbf{Evaluation Metrics} To comprehensively evaluate the effectiveness and efficacy of different pruning methods, we adopt four commonly used metrics as follows. We utilize a) \textbf{FVD}~\cite{PengZZ19}, which is pretained by SVC (StarCraft 2 Videos) dataset, to evaluate the quality of spatial and temporal features in video generation, b) \textbf{Clip-Score}~\cite{HesselHFBC21} to measure the alignment of text and image and coherence of generated videos respectively denoted as Clip(text) and Clip(img), c) \textbf{FLOPS} (floating point operations per second) to assess the floating calculation of pruning techniques, and d) \textbf{time}(s) to provide an assessment of the practical running speed.

\begin{table}[]
\resizebox{\linewidth}{!}{%
\begin{tabular}{@{}l|c|c|c|c@{}}
\toprule
\multicolumn{1}{c|}{Method}          & Clip (img) $\uparrow$ & Clip (text) $\uparrow$ & TFLOPS $\downarrow$ & Time (s) $\downarrow$ \\ \midrule
Tune-A-Video      & 97.78                      & 32.83                       & 379.8               & 45.08             \\ \midrule
+ ToMeSD & 97.73                      & 32.88                       & 369.53              & 48.54             \\
+ F$^3$-Pruning             & \textbf{97.95}             & \textbf{33.30}              & \textbf{355.23}     & \textbf{43.83}    \\ \bottomrule
\end{tabular}
}
\caption{Quantitative comparison results on two pruning methods applied to the typical diffusion-based models Tune-A-Video. Experiment conducts on LONGTEXT. 
%Due to the unavailability of the original reference video, we only utilize the Clip-score to assess the coherence of generated video and text-image alignment, which are respectively denoted by Clip (img) and Clip (text). 
% TFLOPS means Tera FLOPS, $10^{12}$ FLOPS.
}
\label{tab:compDiffuison}
\end{table}

\subsection{Comparison on transformer-based methods}
To verify the efficacy and effectiveness of F$^3$-Pruning, we conduct comparative experiments on UCF-101, DAVIS, and LONGTEXT using the baselines including Token Pruning~\cite{RaoZLLZH21}, Token Reorganization~\cite{MaoXNBFLXX21}, TPS~\cite{WeiTSYJ23}, and ToMe~\cite{BolyaFDZFH23}. We equally prune 50\% for each baseline on TA. The quantitative and qualitative results are respectively shown in Tab.~\ref{tab:compAttention} and Fig.~\ref{transformerComp}.

As indicated in Tab.~\ref{tab:compAttention}, we equally prune 50\% of TA in each baseline and make the FLOPS decline about 44\%. However, only F$^3$-Pruning achieves the lowest inference time, resulting in an approximately 1.35x speedup. On the contrary, other baselines lead to an increase in inference time. It is because the F$^3$-Pruning without additional calculations and tensor merging during the inference stage. For video quality, F$^3$-Pruning exhibits exceptional quality on UCF-101, leading to a significant 22\% improvement in FVD. Similarly, F$^3$-Pruning achieves the best Clip(text) among the five baselines. Although the Clip(text) of ToMe is slightly higher than F$^3$-Pruning, the results of DAVIS dataset also show the same conclusion as UCF-101. To test the performance of F$^3$-Pruning with natural text as input. We conduct experiments on LONGTEXT dataset and prove that F$^3$-Pruning achieves the best Clip(text) among the five baselines. 

The comparison in Fig.~\ref{transformerComp} shows that four competitors lead to misinterpretations of text and the generation of illogical videos. In the case of the CogVideo, it overly connects dense TA to ensure smooth transitions between frames. However, this design might result in reduced motion information within the generated videos. Additionally, incorrect pruning and merging strategies by the other four competitors lead the misleading text and incoherent video. When F$^3$-Pruning prunes temporal redundancy, the model redistributes TA from previous frames to the textual content. Consequently, this adjustment tends to produce coherent videos with a greater emphasis on text motion alignment.

\begin{figure}[t]
	\centering
	\includegraphics[width=0.85\linewidth]{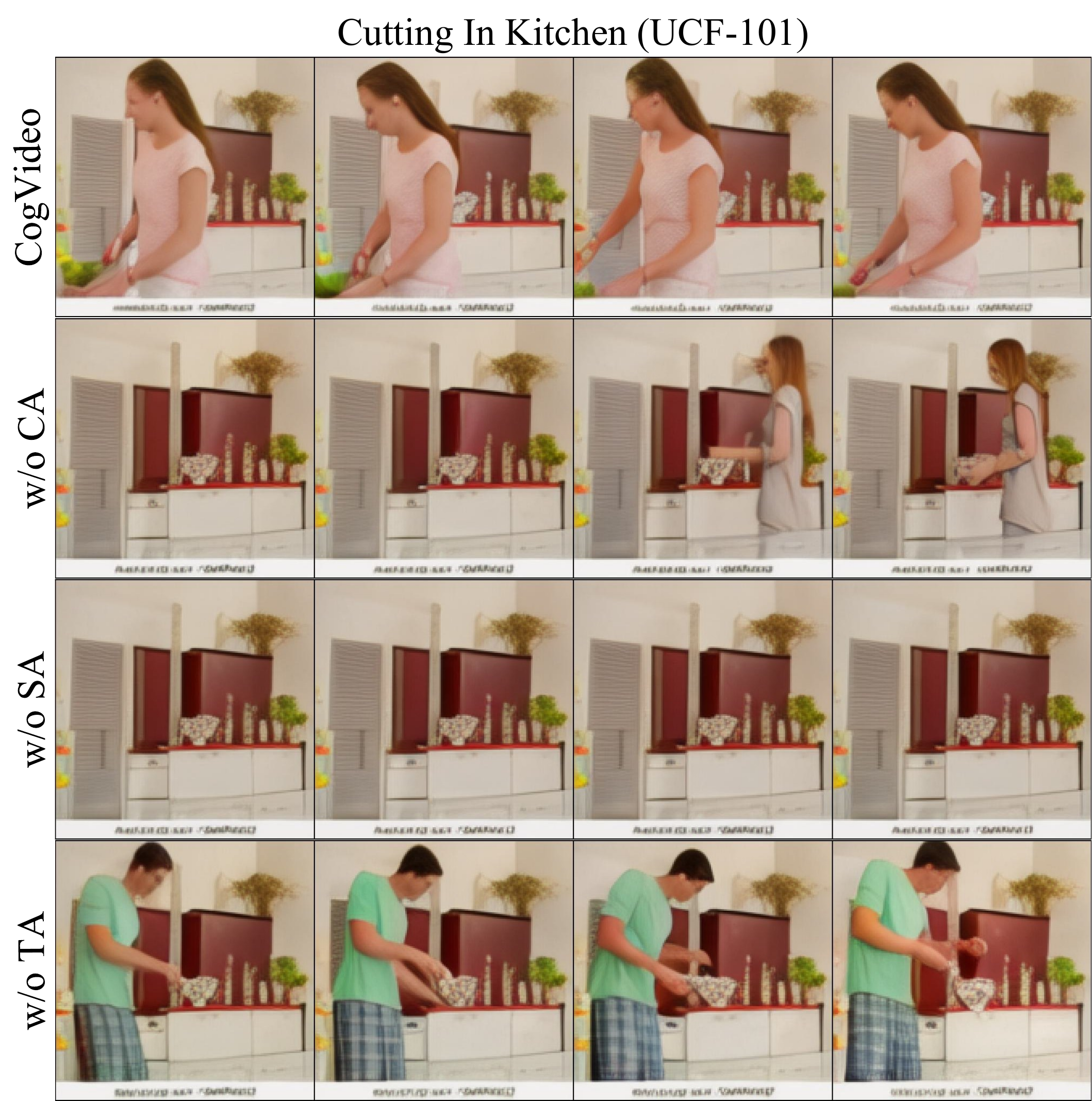}
	\caption{Some examples of pruning TA, SA, and CA.}
	\label{Ablation}
\end{figure}
\begin{figure}[t]
	\centering
	\includegraphics[width=0.85\linewidth]{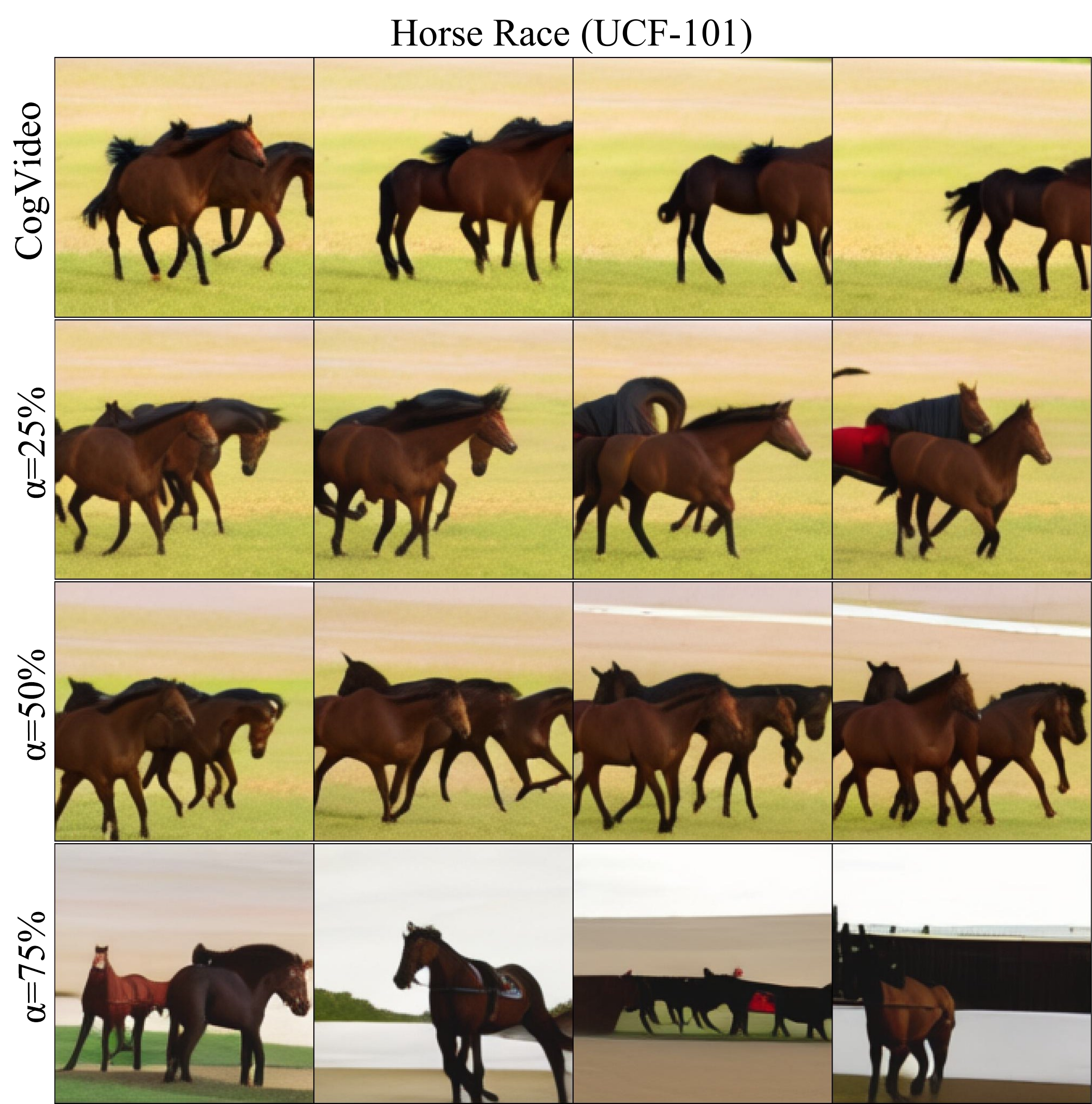}
	\caption{Some examples of different pruning ratios $\alpha$.}
	\label{threshold}
\end{figure}

\subsection{Comparison on the diffusion-based method} \label{Comparison on diffusion-based methods}
To further verify the versatility of F$^3$-Pruning, we conduct experiments on the diffusion-based Tune-A-Video, with the comparison method ToMeSD~\cite{BolyaH23}. As shown in Tab.~\ref{tab:compDiffuison}, F$^3$-Pruning achieves the best overall metrics. Note that, although ToMeSD has a little improvement on FLOPS, the inference time of it is slightly higher than Tune-A-Video. It is because that ToMeSD introduces an extra step of merge and unmerge process, which leads to an impact on the efficiency of generation. Additionally, ToMeSD exhibits a slight decrease in Clip(img) and a minor increase in Clip(text), but F$^3$-Pruning demonstrates a more pronounced improvement in Clip(text) and Clip(img). The visual results in Fig.~\ref{diffusionComp} also show that the excessive attention of Tune-A-Video and ToMeSD to other frames causes overly reliance on preceding frames, resulting in distortions and increased noises in the details. As a result, F$^3$-Pruning is the clear winner in this experiment.

\begin{table}[]
\resizebox{\linewidth}{!}{%
\begin{tabular}{@{}c|c|c|c|c@{}}
\toprule
Methods         & FVD $\downarrow$ & Clip(text) $\uparrow$ & PFLOPS $\downarrow$ & Time(s) $\downarrow$ \\ \midrule
CogVideo       & 5357             & \textbf{25.03}        & 9483                & 486            \\\midrule
w/o CA       & 4730             & 24.97                 & 9446                & 480            \\
% w/o CA      & 4309             & 24.88                 & 6144                & 389            \\
w/o SA       & 5410             & 24.87                 & 8651                & 460            \\
w/o TA & \textbf{4239}    & 24.77                 & \textbf{5311}       & \textbf{359}   \\ \bottomrule
\end{tabular}
}
\caption{Abaltion study of w/o CA, SA and TA on UCF-101.}
\label{tab:abaltion-different-module}
\end{table}

\begin{table}[t]
\resizebox{\linewidth}{!}{%
\begin{tabular}{@{}c|c|c|c|c@{}}
\toprule
Methods & FVD  $\downarrow$  & Clip(text) $\uparrow$ & PFLOPS $\downarrow$ & Time (s) $\downarrow$ \\ \midrule
CogVideo         & 5357                                & \textbf{25.03}                  & 9483                & 486                   \\ \midrule
$\alpha$=25\%      & 4937                                & 24.96                  & 7397                & 422                   \\
$\alpha$=50\%      & 4239                                & 24.77                  & 5310                & 359                   \\
$\alpha$=75\%      & \textbf{1750}                                & 23.52                  & \textbf{3224}                & \textbf{311}                   \\ \bottomrule
\end{tabular}
}
\caption{Ablation Study of pruning ratios $\alpha$ on UCF-101.}
\label{tab:threshold}
\end{table}

\subsection{Ablation Study} \label{Ablation Study}

\textbf{Effect of pruning CA, SA, and TA}
To validate the effects of three attention, we conduct an ablation study on UCF-101 dataset by separately pruning CA, SA, and TA. The results are demonstrated in Tab.~\ref{tab:abaltion-different-module}. Although ``w/o CA" has the improvement of FVD and the lowest harmful for Clip(text),  its inference time has little improvement. Similarly, ``w/o SA" has modest harm for Clip(text), but it increases FVD, thereby reducing the quality of generated video.  Differently, ``w/o CA'' significantly reduced inference time, and making the best trade-off between costs and quality among those three. Our visual results, depicted in Fig.~\ref{Ablation}, underscore the consequence, revealing ``w/o CA" and ``w/o SA" have a loss of alignment from textual input. As a result, we choose the TA as our pruning attention.

\textbf{Effect of different pruning ratios}
To determine the optimal pruning ratio $\alpha$, we conduct an ablation study to prune CogVideo by 25\%, 50\%, and 75\% on UCF-101. Qualitative results are demonstrated in Tab.~\ref{tab:threshold}). Notably, as the pruning ratio increases, we observe a huge reduction in inference time and FLOPS. These there pruning ratios lead to a substantial enhancement in FVD. And 25\% and 50\% exhibit fewer effects on the final Clip(text), suggesting they maintain a robust text-image alignment. However, the experiment of 75\% pruning ratio achieves the best FVD performance and inference time, but it has a notable decline in text-video alignment. This observation is further supported by the findings presented in Fig.~\ref{threshold}. As shown in the figure, a 50\% pruning ratio demonstrates modest improvements in text-image alignment and inter-frame coherence within the generated videos. In contrast, the experiment with a 75\% pruning ratio exhibits a significant loss of coherence among frames. As a result, we select a 50\% pruning ratio for all our experiments.

\section{Conclusion}
In this paper, to accelerate inference of T2V, we propose a training-free and generalized pruning strategy called F$^3$-Pruning. Specifically, temporal attention weights will be pruned if their aggregate attention values rank below a pruning ratio.
Experiments prove that our method promotes inference speed and video quality on mainstream T2V models.

\bibliography{aaai24}

\newpage
\begin{appendices}

\begin{table}[b]
\centering
\begin{tabular}{@{}c|ccc@{}}
\toprule
Datasets & Text inputs & Batch size & Total videos \\ \midrule
UCF-101  & 101   & 10         & 1010         \\
DAVIS    & 60    & 8          & 480          \\
LONGTEXT & 50    & 8          & 400          \\ \bottomrule
\end{tabular}
\caption{Details of datasets.}
\label{tab:implementation-details}
\end{table}

\section{Implementation Details}
%  Since our method primarily focuses on addressing cross-frame redundancy during video generation, we specifically select the first stage of CogVideo for ablation, which is dedicated to video generation, as the test bed for evaluating our approach on Transformer-based models.
% We conduct experiments on the CogVideo (Transformer-based) and Tuneavideo (Diffusion-based). In our experiments on CogVideo, we utilize the UCF-101 label as the input. We generate 10 videos for each label, resulting in a total of 100 videos. For the DAVIS experiment, we generate 16 videos per tag in two times, amounting to a total of 960 videos. In the LONGTEXT experiments, we generate 8 videos for each long sentence, totaling 400 videos. And in the Tuneavideo experiment, we employ examples as input to edit videos on LONGTEXT dataset. To assess the quality and speed of generation, we compare our results with the SOTA model, Token Pruning~\cite{RaoZLLZH21}, Token Reorganization~\cite{MaoXNBFLXX21}, TPS~\cite{WeiTSYJ23}, and tomesd~\cite{BolyaH23}. It is important to note that the forward stage of the i3d model in FVD has a minimum frame count limit of $10$. Since CogVideo 1-stage only generates $5$ frames, we concatenate two $5$-frame of different videos to create a $10$-frame video. However, for full-stage CogVideo, which generates $33$ frames, we discard the first frame and split the remaining frames into two $16$-frame videos for evaluation. All experiments are conducted on NVIDIA A6000 GPUs. 
There are two stages for CogVideo. The first stage focuses on the generation of key frames. The second stage focuses on interpolation and super-resolution of the key frames. Since F$^3$-Pruning primarily addresses temporal redundancy during video generation, we specifically select the first stage of CogVideo for the ablation study. 

% It is noteworthy that the forward stage of the i3d model in FVD has a minimum frame count of $10$. Since the first stage of CogVideo only generates $5$ frames, we concatenate the two $5$ frames of different videos with the same label to create a video of $10$ frames. However, for both-stage CogVideo, which generates $33$ frames, we discard the first frame and split the remaining frames into two $16$-frame videos for evaluation. 

Some details of datasets are described in Tab.~\ref{tab:implementation-details}. All experiments are conducted on NVIDIA A6000 GPUs. 

\begin{figure*}[t]
	\centering
	\includegraphics[width=1\linewidth]{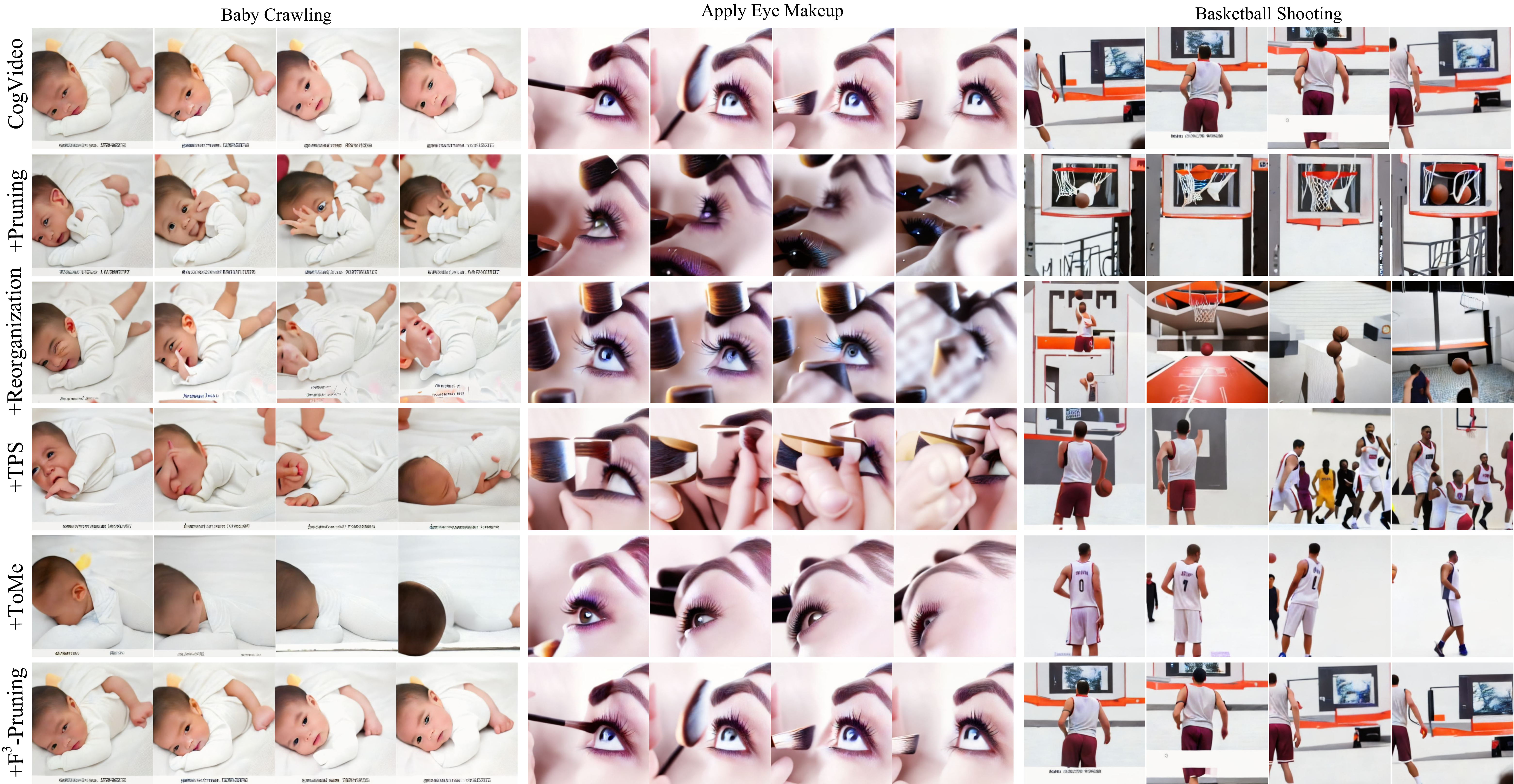}
	\caption{More visual results of different pruning methods on the dataset of UCF-101.}
	\label{appTransformerUCF}
\end{figure*}

\begin{figure*}[t]
	\centering
	\includegraphics[width=1\linewidth]{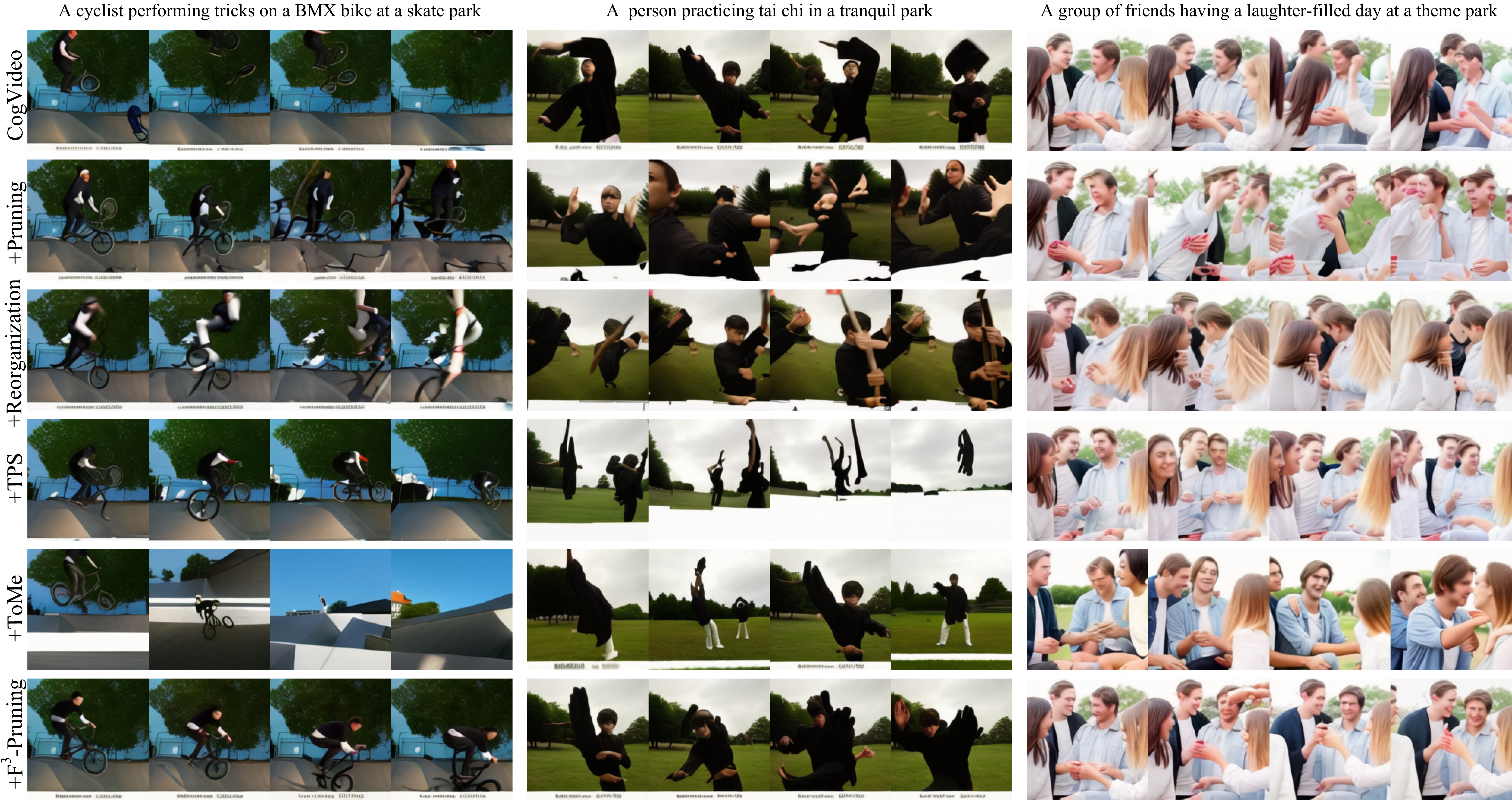}
	\caption{More visual results of different pruning methods on the dataset of LONGTEXT.}
	\label{appTransformerLONG}
\end{figure*}

\begin{figure*}[]
	\centering
	\includegraphics[width=1\linewidth]{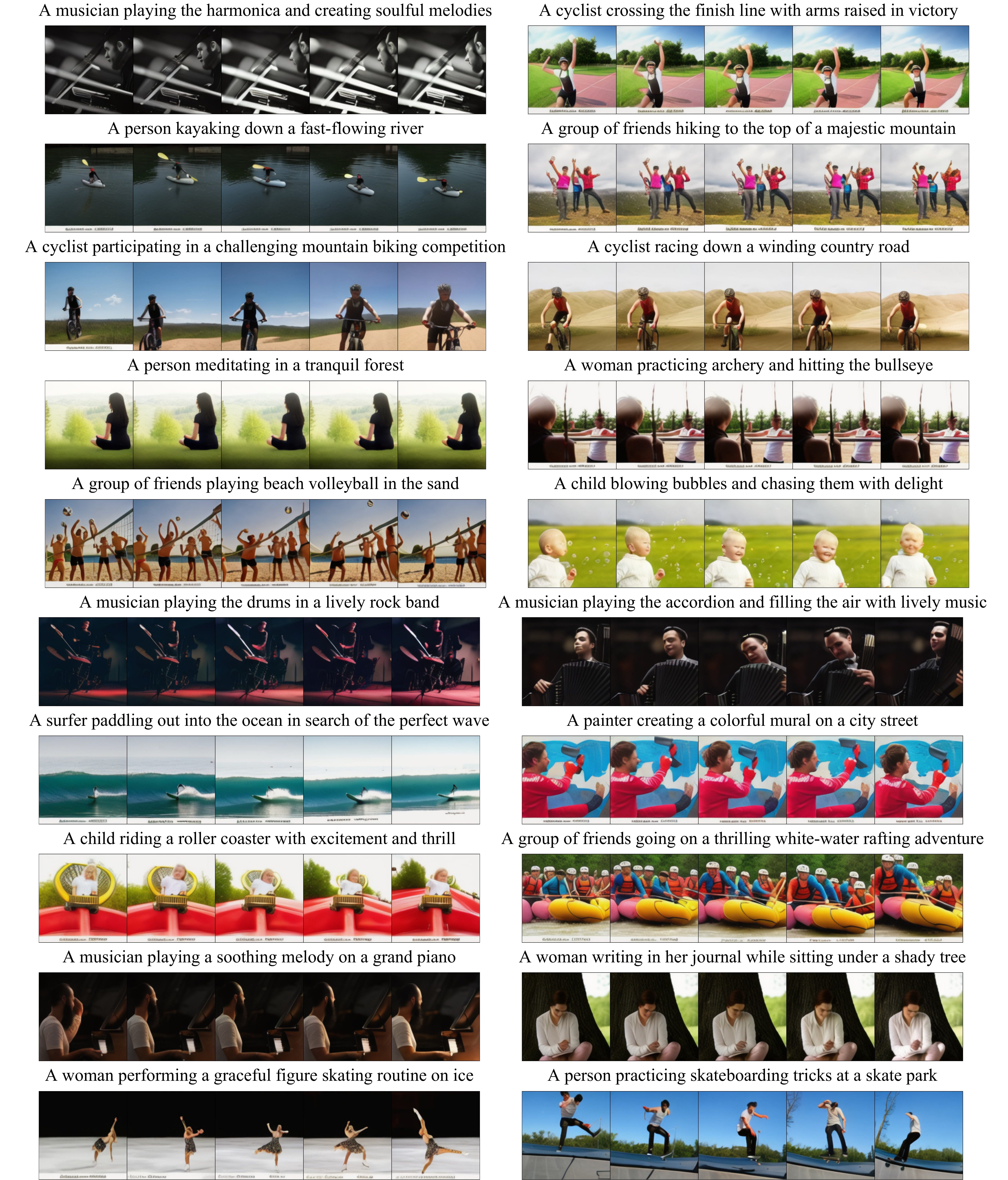}
	\caption{More visual results of F$^3$-Pruning applied to CogVideo on the dataset of LONGTEXT.}
	\label{app-AA-Pruning}
\end{figure*}

\section{More Visual Results}
To provide additional support for the effectiveness of F$^3$-Pruning, we respectively show more comparison visual results in Fig.~\ref{appTransformerUCF} and Fig.~\ref{appTransformerLONG} on UCF-101 and LONGTEXT using CogVideo. Both results show that four competitors generate incoherent and less text-aligned videos. For example, they generate distorted faces and different people in the adjacent frames. And F$^3$-Pruning avoids the above problems to a certain extent and shows the robust text-visual alignment and video coherence as the original CogVideo. And more visual results of F$^3$-Pruning applied to CogVideo on LONGTEXT are shown in Fig.~\ref{app-AA-Pruning}.

\end{appendices}

\end{document}